\documentclass[runningheads]{llncs}
\usepackage[T1]{fontenc}
\usepackage{graphicx}
\usepackage{mathtools} 
\usepackage{booktabs} 
\usepackage{tikz} 
\usepackage{amsmath}
\usepackage{listings}
\usepackage{multirow}

\begin{document}
\title{OR-Agent: Bridging Evolutionary Search and Structured Research for Automated Heuristic Design}
%
%
\author{Qi Liu\inst{1}\orcidID{0000-0002-1587-8524} \and
Ruochen Hao\inst{2}\orcidID{0000-0002-1162-1879} \and
Can Li\inst{2}\orcidID{0009-0006-3544-9336} \and
Wanjing Ma\inst{2}\orcidID{0000-0002-9403-3174}}

\authorrunning{Q. Liu et al.}
%
\institute{School of Transportation, Jilin University, Changchun, 130022, China
\email{liuqijlu@jlu.edu.cn} \and
Key Laboratory of Road and Traffic Engineering of the Ministry of Education, College of Transportation, Tongji University, Shanghai, 201804, China \\
\email{\{haoruochen, lican, mawanjing\}@tongji.edu.cn}}
\maketitle              
\begin{abstract}
Automating heuristic design in complex, experiment-driven domains requires more than iterative mutation of solution algorithms. Current LLM-based evolutionary methods often rely on stochastic mutation loops that lack long-term strategic planning and a formal mechanism to learn from historical failures, leading to inefficient exploration and redundant trials. To address this, we present OR-Agent, a multi-agent research framework designed for automated heuristic design in optimization problems with rich experimental environments. OR-Agent organizes heuristic search as tree-based workflow that explicitly models branching hypothesis generation and systematic backtracking. Furthermore, to address the lack of adaptive learning in current agents, we introduce a hierarchical, optimization-inspired reflection system in which short-term reflections act as verbal gradients, long-term reflections as verbal momentum, and memory compression as semantic weight decay — collectively forming a principled mechanism for governing research dynamics. Extensive experiments on classical combinatorial optimization problems (e.g., TSP, CVRP, bin packing) and simulation-based cooperative driving scenarios demonstrate that OR-Agent outperforms strong evolutionary search baselines. All code and experimental data are publicly available at \url{https://github.com/qiliuchn/OR-Agent}.

\keywords{Automated Heuristic Design \and Evolutionary Search \and Combinatorial Optimization.}
\end{abstract}

\section{Introduction}

Scientific research is traditionally an iterative and human-driven process. Researchers survey existing literature, formulate hypotheses, design algorithms or models, conduct experiments, analyze outcomes, and refine ideas through repeated feedback cycles. This workflow is rarely linear: it involves branching exploration, backtracking from unproductive directions, local refinement around promising ideas, and continual integration of new insights~\cite{tang2025ai}.

Recent advances in large language models (LLMs) and autonomous agents \cite{gottweis2025towards} have raised an increasingly compelling question: Can AI systems act as research collaborators, supporting or partially automating the scientific discovery process? Emerging work has demonstrated that LLMs can generate code, propose hypotheses, perform reasoning, and even participate in iterative refinement~\cite{romera2024mathematical}. However, most existing approaches focus either on single-shot program generation or on evolutionary-style search with limited structured memory and workflow management \cite{romera2024mathematical,liu2023algorithm,liu2024evolution,ye2024reevo,novikov2025alphaevolve}. A comprehensive framework that integrates idea generation, implementation, experimentation, reflection, and structured exploration remains underdeveloped.

In this paper, we introduce \emph{Open Research Agent (OR-Agent)}, a multi-agent system designed to emulate and systematize the workflow of scientific research, with a particular focus on automatic heuristic design for operations research (OR) problems. OR-Agent treats heuristic ideas and programs as research artifacts to be iteratively evolved, evaluated, and refined \cite{liu2024evolution}. Rather than relying solely on mutation or crossover style search iteration \cite{romera2024mathematical,liu2023algorithm,liu2024evolution,novikov2025alphaevolve}, OR-Agent organizes the research process as a structured tree of investigations. The system integrates evolutionary initialization, deep local investigation, and memory-based reflection to balance exploration and exploitation under finite computational budgets. 

We conduct extensive experiments across classical combinatorial optimization problems (e.g. TSP, CVRP, bin packing) and simulation-based cooperative driving scenarios. Results demonstrate OR-Agent’s capability to outperform baseline algorithms across diverse problem domains.

\section{Open Research Agent}

Each research process begins with a human-in-the-loop problem specification stage. OR-Agent provides an \emph{Open Research Canvas (OR-Canvas)} as a shared interface for human researchers and AI agents to collaboratively define the research objective, evaluation protocol, and seed heuristic.

\subsection{OR-Agent Framework}

\begin{figure}
    \centering
    \includegraphics[width=1.0\linewidth]{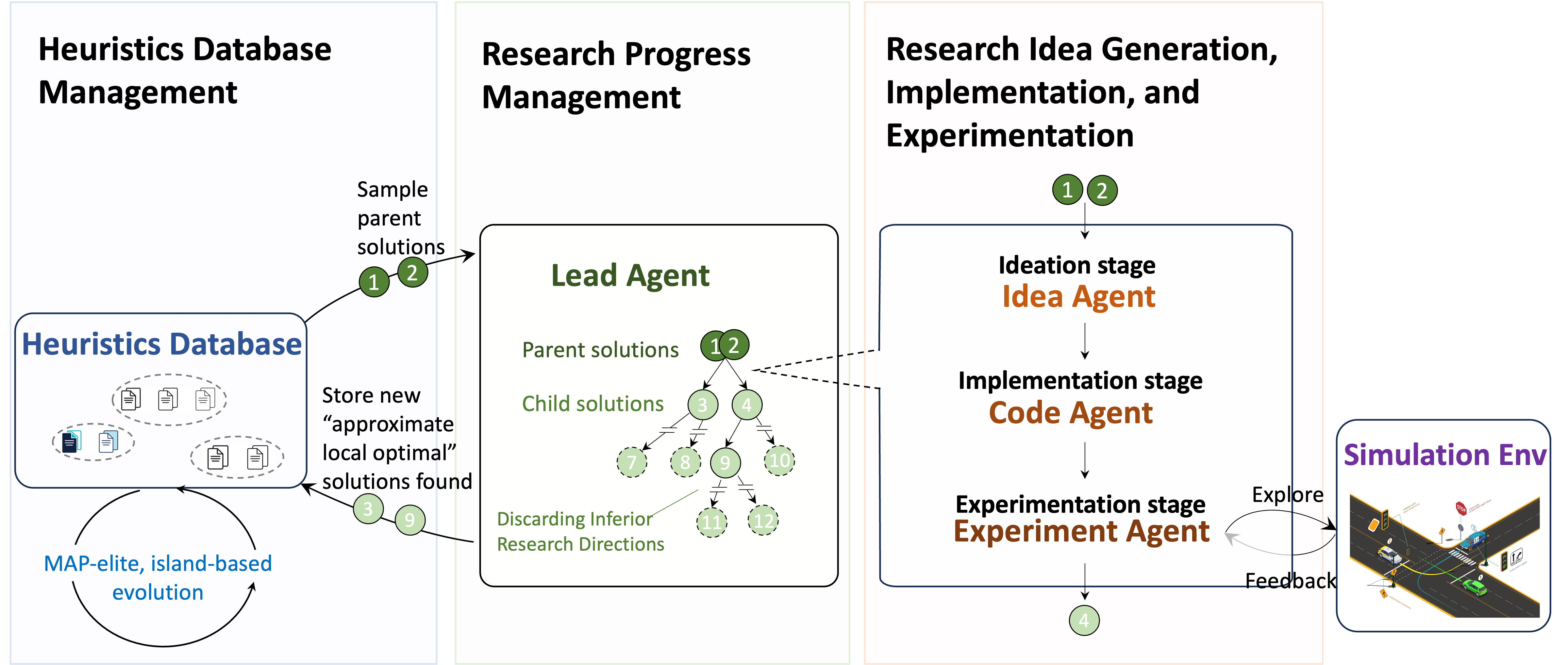}
    \caption{Overall framework of OR-Agent, illustrating the interaction between evolutionary initialization, multi-agent research workflows, experimentation, reflection, and the shared heuristics database.}
    \label{fig:framework}
\end{figure}

OR-Agent distributes exploration across specialized agents while centralizing knowledge through a shared heuristics database and reflections, achieving both research breadth and depth. OR-Agent is implemented as a coordinated multi-agent system with clearly separated responsibilities (Figure \ref{fig:framework}):
\begin{itemize}
    \item \textbf{OR Agent}: The system entry point that manages global configuration, coordinates multiple Lead Agents, and maintains the shared heuristics database.
    \item \textbf{heuristics database}: A persistent repository that stores all generated solutions and associated metadata, serving as a shared evolutionary memory across agents and research rounds.
    \item \textbf{Lead Agent}: Acts as a principal investigator for a research round, selecting parent heuristics, managing the research workflow, and deciding when to expand, backtrack, or terminate exploration.
    \item \textbf{Idea Agent}: Generates and refines high-level solution ideas based on prior experiments, accumulated reflections and current research tree progress.
    \item \textbf{Code Agent}: Translates ideas into algorithm implementations and performs iterative debugging and refinement.
    \item \textbf{Experiment Agent}: Executes experiments, explores complex environments, diagnoses failures, and summarizes experimental findings.
\end{itemize}

At the start of each research round, a \emph{Lead Agent} samples one or more parent heuristics from a structured heuristics database, analogous to population initialization in evolutionary algorithms. Sampling strategies range from near-uniform selection (favoring exploration) to elite-biased selection (favoring exploitation), with temperature-controlled interpolation between the two. Mutation- and crossover-like operations are realized through LLM–driven idea recombination and variation, similar in spirit to prior LLM-based genetic approaches such as AEL~\cite{liu2023algorithm}, EoH~\cite{liu2024evolution} and ReEvo~\cite{ye2024reevo}. However, unlike these methods, OR-Agent does not rely on frequent evolutionary operators alone. Instead, it emphasizes extensive and systematic investigation around each evolutionary starting point before progressing to new ones. This design choice reflects a key departure from prior work: in scientific research, generating ideas via mutation or crossover is only the beginning. High-quality discoveries typically emerge from iterative refinement, targeted experimentation, and repeated diagnosis of failure modes. OR-Agent explicitly allocates computational effort toward deep local exploration once a promising direction is identified, closely mirroring human research practice. Each Lead Agent organizes its research process as a tree-structured workflow. Multiple Lead Agents may operate concurrently, enabling parallel exploration of distinct research directions while sharing knowledge through the common heuristics database.

\subsection{Evolutionary Heuristics Database}
OR-Agent maintains a shared structured evolutionary heuristics database that stores previously evaluated programs together with rich evaluation feedback and experiment summaries. The design is inspired by the program database paradigm used in LLM-driven evolutionary systems such as FunSearch~\cite{romera2024mathematical} and AlphaEvolve\cite{novikov2025alphaevolve}. A key motivation for maintaining a persistent evolutionary database is an empirically observed failure mode, which we term ``population ruin''. Without an explicit mechanism to preserve diversity and prevent over-concentration, a single seemingly ``good'' solution can rapidly dominate the population, even if it is brittle or invalid under refinement. In practice, such dominance is dangerous because the dominating solution may contain degenerate behaviors (e.g., excessive computation leading to timeouts or invalid outputs), and subsequent mutations around this solution can wipe out the remaining viable population.

\subsection{Tree-Structured Scientific Discovery Workflow}

Deep scientific research is inherently challenging, requiring both broad exploration over diverse high-level approaches and deep, sustained refinement within promising directions. OR-Agent explicitly maintains the structural relationships among hypotheses, allowing deeper investigation along promising branches while preserving alternative directions for later exploration. Crucially, during idea generation, OR-Agent incorporates the entire research tree—including node ideas, evaluation scores, and expansion status—into the generation context. This global awareness provides the agent with a holistic view of research progress, analogous to how a human scientist tracks explored directions, failed attempts, and promising ideas over time.

\begin{figure}
    \centering
    \includegraphics[width=0.7\linewidth]{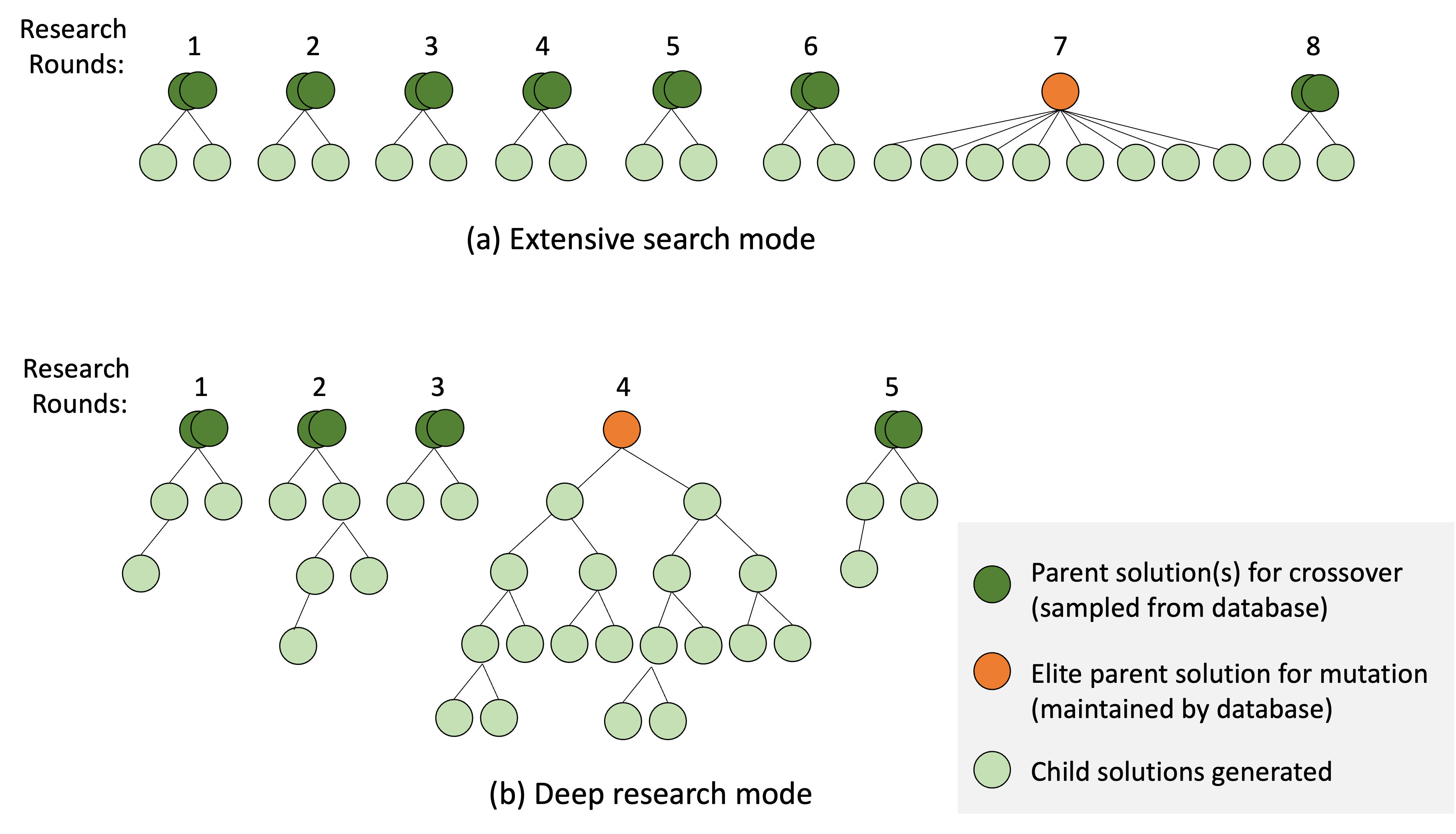}
    \caption{Use tree shape as a control mechanism for research modes illustration.}
    \label{fig:workflow_modes}
\end{figure}

\paragraph{Tree Shape as a Control Mechanism for Research Modes.}
The depth and breadth of the research tree naturally correspond to the depth and breadth of scientific inquiry. Deeper trees emphasize intensive refinement of specific ideas, while wider trees favor exploration of diverse hypotheses. OR-Agent exposes multiple parameters to explicitly control tree shape, including the maximum number of children per node, maximum tree depth, and dynamic child allocation strategies. Through these controls, OR-Agent supports various research modes, allowing users to trade off exploration and exploitation. Representative modes such as ``deep research'' and ``extensive search'' are illustrated in Figure~\ref{fig:workflow_modes}.

While related systems such as AI Scientist V2\cite{yamada2025ai} and FlowSearch \cite{hu2025flowsearch} also employ tree- or graph-based workflows, their structures are designed to support more general task decomposition and knowledge propagation. In contrast, the tree-structured research workflow in OR-Agent is specifically tailored to scientific idea development, where edges encode natural derivation relationships between research ideas.

\paragraph{Coordinated Idea Generation.}
The ability to generate innovative ideas is crucial. Prior studies have observed that naively scaling LLM-based idea generation often leads to severe redundancy and limited diversity \cite{si2024can}. OR-Agent addresses this limitation through \emph{coordinated idea generation}. For each node expansion, the Idea Agent generates a set of child ideas jointly, conditioning on the shared parent context or even entire research tree and the requirement that the ideas represent distinct directions. This process treats the set of children as a coherent research plan rather than independent samples.

The prompt used by the Idea Agent is shown in Listing~\ref{lst:idea_agent_prompt}. Without coordinated idea generation, the diversity of generated ideas remains surprisingly limited, with most ideas exhibiting nearly identical structural patterns, as shown by an example from the Online Bin Packing Problem (Listing~\ref{lst:independent_generation_bpp}). In contrast, under coordinated idea generation, the Idea Agent produces substantially more diverse ideas, as illustrated in Listing~\ref{lst:coordinated_idea_generation_bpp}.

We note that coordinated generation also introduces practical trade-offs. Generating multiple ideas in a single pass tends to produce much shorter individual descriptions, and the total number of ideas is constrained by context window limits.

\begin{lstlisting}[caption={Prompt used by Idea Agent to generate research ideas.},
label={lst:idea_agent_prompt}]
## Context
Problem description: {problem_description}
Name of the function to optimize: `{function_to_evolve}`.
The goal is to {obj_type} the `score` of `{function_to_evolve}` evaluation result.
Function signature requirement: {function_description}
Current Research Flow Graph: {current_research_flow_graph}
Parent solution(s) you can learn from: {parent_solutions}
Current long-term research reflection: {long_term_reflection}

## Task
Based on the analysis of the parent solutions, the current research flow graph, and the long-term research reflection (if available), generate innovative ideas for improving the `{function_to_evolve}` function.

### Research Planning Approach
Think like a research planner organizing a scientific investigation. Generate ideas that:
- **Explore different research directions**: Propose ideas that investigate various aspects of the problem; detailed plan is preferred!
- **Consider the research flow graph**: Use the current research flow graph to identify gaps and complement ongoing investigations
- **Can be complementary**: Ideas can work together to form a comprehensive research plan
- **Are independently implementable**: Each idea can be developed and tested separately
- **Address different improvement areas**: Cover multiple potential solution strategies

You may generate up to {num_ideas} specific, actionable ideas that collectively form a comprehensive research plan.
\end{lstlisting}

\begin{lstlisting}[caption={Independently generated child ideas for BPP-Online from the same parent solution.}, 
label={lst:independent_generation_bpp}]
1. Combine value-to-weight ratio with a sparsification mechanism that zeros out items whose total weight exceeds a dynamic threshold based on average knapsack capacity usage. Specifically, compute the initial heuristic as prize divided by the L1 norm of weight (as in the elite solution), then set to zero any item whose max weight across constraints exceeds 1/sqrt(m) (reflecting tighter per-dimension feasibility), or alternatively, zero out the bottom 30% of items by this ratio to focus search on most promising candidates.
2. Combine value-to-weight ratio with a sparsification mechanism that zeros out items whose total weight exceeds the average capacity usage per item, to focus selection on both efficient and feasible items. Specifically, compute the initial heuristic as prize divided by the L1 norm of weight (as in the elite solution), then set to zero any item whose total weight (sum across constraints) is greater than the median total weight of all items, thereby promoting inclusion of lighter, high-value items while pruning potentially infeasible heavy ones early.
3. Combine value-to-weight ratio with a sparsification mechanism that zeros out items whose total weight exceeds a dynamic threshold based on average knapsack capacity usage. Specifically, compute the initial heuristic as prize divided by the sum of weights (as in the elite solution), then set to zero any item whose maximum weight across all constraints exceeds 1 / m (the average per-constraint capacity), making it less likely to be selected due to high risk of violating constraints.
4. Combine value-to-weight ratio with a sparsification mechanism that zeros out items whose total weight exceeds the average capacity usage per item. Specifically, compute the standard value-to-total-weight ratio (prize / sum(weight, axis=1)), then identify items with total weight greater than the median total weight across all items; set their heuristic value to zero to discourage selection of overly heavy items unless they are exceptionally valuable. This balances desirability with feasibility under multiple constraints.
5. Combine value-to-weight ratio with a sparsification mechanism that zeros out items whose total weight exceeds the average capacity pressure across constraints. Specifically, compute the initial heuristic as prize divided by the L1 norm of weight (as in the elite solution), then set to zero any item where the maximum weight across constraints exceeds a threshold (e.g., 1/m or the median of max weights). This focuses ant attention on items that are both valuable per unit resource and feasible under tightest constraints.
...(Check public repository for the complete list)...
\end{lstlisting}

\begin{lstlisting}[caption={Coordinated child ideas for BPP-Online generated in a single LLM call.}, 
label={lst:coordinated_idea_generation_bpp}]
1. Normalize the prize-to-weight ratio per constraint dimension and aggregate via geometric mean to better reflect balanced resource usage across all constraints.
2. Introduce a sparsification step that sets heuristic values to zero for items whose total weight exceeds a threshold (e.g., average capacity usage), focusing selection on feasible candidates.
3. Incorporate a penalty term based on the maximum weight of an item across all constraints to discourage selection of items that dominate any single resource.
4. Compute heuristic as prize divided by the L2 norm of the weight vector, emphasizing compactness in multi-dimensional weight space.
5. Use min-max normalization of weights per dimension before computing prize-to-weight ratios to ensure equitable contribution from each constraint.
...(Check public repository for the complete list)...
\end{lstlisting}

\paragraph{Research Tree Traversal.}
The Lead Agent is responsible for traversing and expanding the research tree. At each step, Lead Agent selects a leaf node to expand according to a simple greedy policy: among all unfinished leaf nodes, the one with the best evaluation score is selected as the current most promising research direction. This choice reflects a strong exploitation bias, though alternative traversal strategies—such as PUCB-style exploration \cite{rosin2011multi} are compatible with the framework and left for future investigation. The tree traversal and expansion procedure for a single research round is summarized in Listing~\ref{lst:research_tree_traversal}:

\begin{lstlisting}[caption={OR-Agent research tree traversal procedure.},
label={lst:research_tree_traversal}]
Initialization:
- Sample parent heuristics from the heuristics database to form the root node.

Research Start:
- Expand the root node to generate up to `max_children` children nodes.

Research Loop:
1. Select the best unfinished leaf node N_i.
2. Expand N_i by generating child ideas, code implementations, and experimental results.
3. Truncation and local optimality test:
   - Retain only children that improve upon N_i.
   - If no child improves upon N_i, mark N_i as terminal.
4. Terminate when all leaf nodes are terminal; otherwise repeat.
\end{lstlisting}

\begin{figure}
    \centering
    \includegraphics[width=1.0\linewidth]{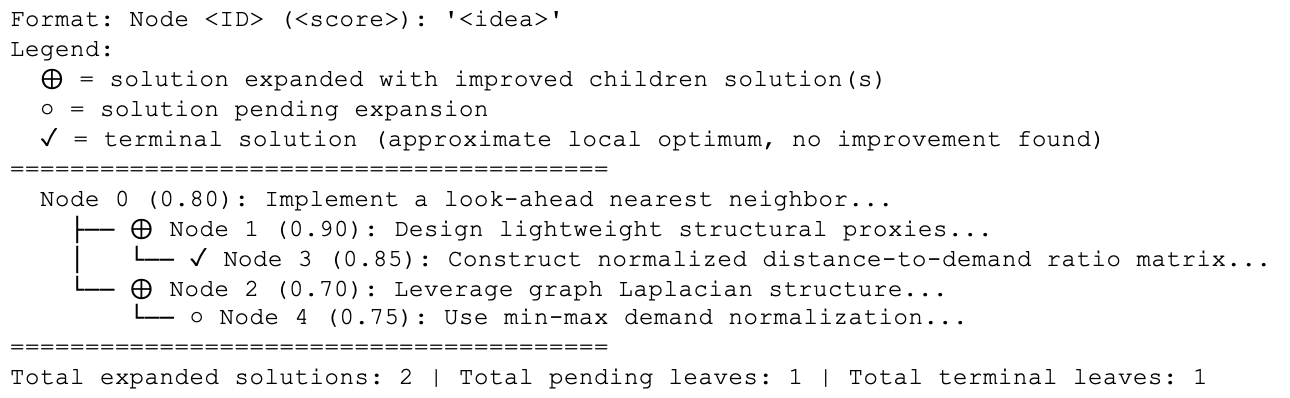}
    \caption{Research tree representation used by OR-Agent.}
    \label{fig:research_tree_representation}
\end{figure}

At the start of one research round, OR-Agent allows temporary performance degradation when exploring new research directions. Child nodes are not required to outperform the root or previous elite immediately, reflecting the reality that early experiments along a novel direction may initially underperform. Termination decisions are instead governed by sustained lack of improvement. The serialization format of research tree used by OR-Agent to is illustrated in Figure~\ref{fig:research_tree_representation}.

\begin{figure}
    \centering
    \includegraphics[width=1.0\linewidth]{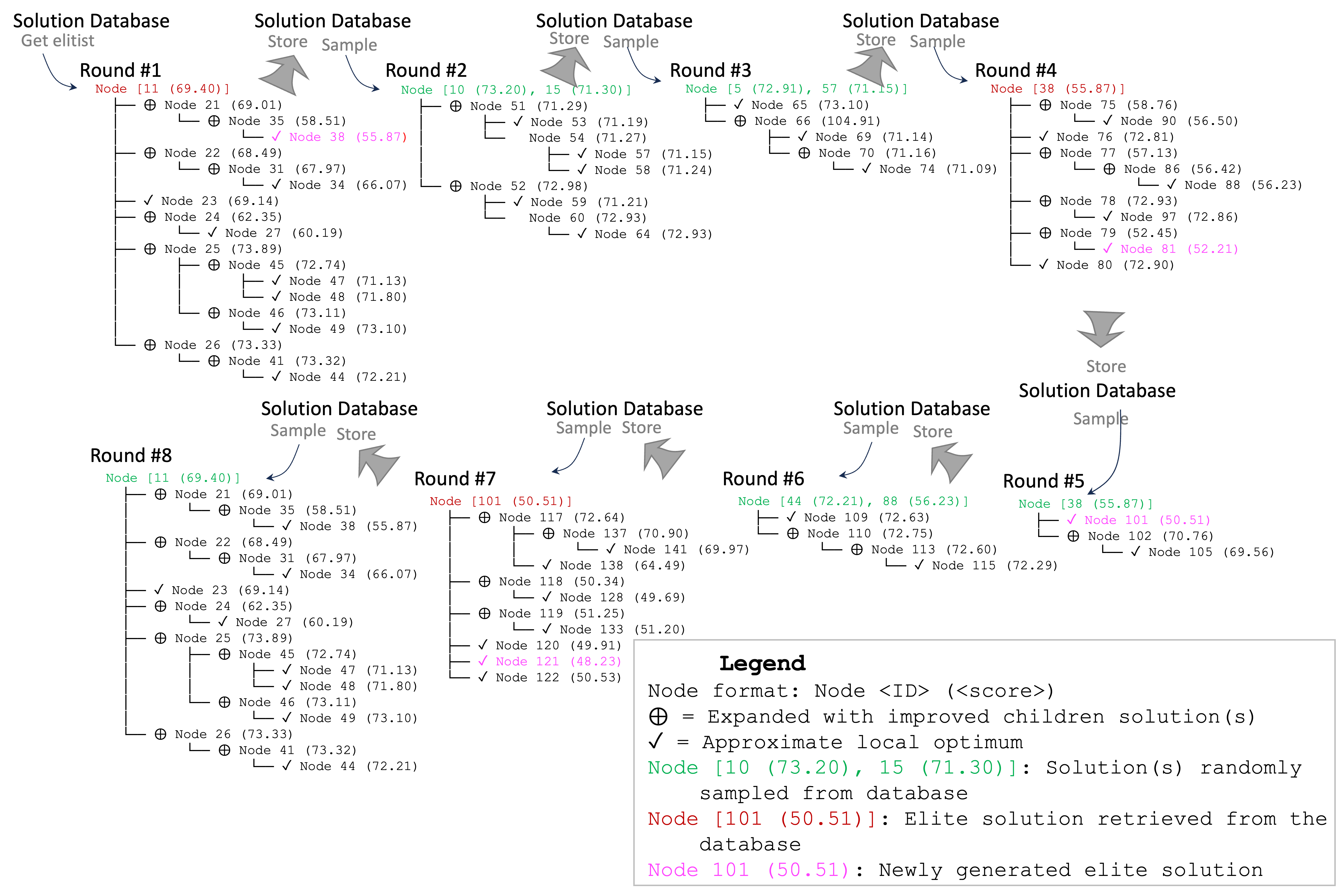}
    \caption{Illustration of dynamic research tree evolution across multiple research rounds.}
    \label{fig:research_rounds}
\end{figure}

Figure~\ref{fig:research_rounds} provides an illustrative example of multiple research rounds, showing how the size and shape of the research tree evolve dynamically over time. Promising research directions are explored more deeply, while unproductive branches are pruned. Although many high-quality solutions originate from elite parent nodes, the figure also highlights that valuable discoveries can emerge from non-elite starting points, underscoring the importance of maintaining sufficient exploratory breadth.

\subsection{Iterative Experimentation and Environment Probing}

OR-Agent incorporates an \emph{Experiment Agent} to iteratively evaluate and refine each candidate solution node in an environment-aware manner. The Experiment Agent operates within a closed-loop workflow that identifies execution failures and performance bottlenecks, performs targeted modifications to the code or parameter settings, interacts with the environment through callbacks when additional information is required, and finally generates a concise experimental report summarizing the key findings and insights to support the development of improved ideas in subsequent iterations.

\paragraph{Environment Exploration via Callbacks.}
In complex environments, aggregate metrics alone are frequently insufficient to diagnose failure causes (e.g., timeouts, deadlocks, safety violations, or brittle corner cases). OR-Agent therefore augments experimentation with environment exploration through callbacks. The Experiment Agent can request changes to callback definitions that control what the environment exposes. By iteratively refining callbacks, the agent can progressively narrow down hypotheses about failure mechanisms, closely mirroring the scientific practice of instrumenting an experiment to observe previously hidden variables.

\paragraph{Dynamic Allocation of Experiment Rounds.}
Using a fixed number of experiment rounds for every candidate is often an inefficient use of computational resources. Candidates with strong potential typically benefit from additional refinement, whereas those showing little promise should be discarded early. To address this issue, OR-Agent implements \emph{dynamic experiment round allocation} strategy, where the maximum number of refinement attempts for a candidate is adapted based on its observed potential, such as its improvement over the parent solution or its elitist status. In general, candidates demonstrating consistent progress are granted additional experimentation opportunities, while those with deteriorating performance receive a reduced computational budget.

\paragraph{Context Compression for Extended Experiment Histories.}
To support long-running refinement processes without exceeding the language model’s context window, OR-Agent periodically compresses the accumulated experiment history into a progressively updated summary. This condensed report retains the most important information from previous iterations, including the best performance achieved, overall performance trends, major observations, and unresolved issues, allowing experimentation to continue efficiently even over many refinement rounds.

\paragraph{Experiment Summarization and Knowledge Extraction.}
After the experimental phase is completed, the Experiment Agent is responsible for generating a comprehensive report that synthesizes the outcomes of all trials. The report summarizes key observations, including overall performance trends, the effects of different parameter settings, recurring failure cases and negative findings, factors that consistently contributed to improved performance, and outstanding challenges that may require fundamental redesign. This consolidated report serves as distilled experimental evidence and is attached to the corresponding solution node, providing the Lead Agent and the Idea Agent with valuable guidance for formulating more effective hypotheses in subsequent iterations.

\paragraph{Learning from Failures and Invalid Outcomes.}
In complex optimization tasks, candidate solutions often exhibit partial success, such as achieving performance gains on small problem instances while failing or timing out on larger ones. Rather than dismissing these outcomes as unsuccessful, OR-Agent leverages them as informative feedback for future refinement. The Experiment Agent analyzes failed or invalid executions to identify useful observations and extract actionable insights that can guide subsequent iterations.

\paragraph{Solution Reversion Mechanisms.}
Iterative refinement can regress performance. OR-Agent supports two complementary reversion strategies. \emph{Soft reversion} encourages the Experiment Agent—through explicit prompting near the final attempt—to revert to a previously better configuration if recent changes degrade performance. Because soft reversion relies on the model’s compliance, OR-Agent additionally implements \emph{hard reversion}: the experimentation controller tracks the best-performing code snapshot during the experiment sequence and restores it upon termination when the final version is worse. The final experiment summary explicitly records the reversion decision and the rationale, ensuring that the downstream workflow reasons over the best available implementation while retaining diagnostic knowledge from the unsuccessful variants.

\subsection{Reflection Mechanism as Semantic Optimizer}

Reflection has been identified as a critical capability for autonomous agents to improve performance through iterative self-evaluation and revision \cite{shinn2023reflexion}. The central view adopted in this work is that \emph{designing a reflection mechanism is analogous to designing an optimizer in conventional numerical optimization}.

OR-Agent implements a hierarchical reflection stack that aggregates feedback at increasing temporal scales, similar to how optimization algorithms accumulate information over minibatches, epochs, and training trajectories. Prior work has described short-term reflection as a form of \emph{verbal gradient} \cite{ye2024reevo}, where natural-language feedback functions analogously to gradient information in numerical optimization. OR-Agent generalizes this perspective by mapping multiple levels of reflection to distinct components of optimization dynamics. 

Step-level experiment reflections operate as local verbal gradients, encoding immediate corrective directions derived from the most recent environmental feedback, and guiding subsequent micro-updates to code or parameters. Aggregated experiment summaries act as batch-averaged verbal gradients, integrating signals across multiple trials to reduce variance, emphasize stable causal effects, and distinguish transient fluctuations from persistent improvements. At a higher level, long-term reflections function as semantic momentum, accumulating recurring lessons across solution trajectories to stabilize search. Additionally, reflection compression introduces an implicit regularization mechanism. It resembles exponential decay in optimizer state, where older or weaker signals diminish unless continually reinforced. Collectively, these mechanisms constitute a principled hierarchical reflection system that governs research dynamics in OR-Agent.

\subsection{Balancing Exploration and Exploitation}

Balancing exploration and exploitation has long been a fundamental challenge in optimization and search. Conventional genetic algorithms address this trade-off implicitly through population evolution, selection mechanisms, and genetic operators. However, these mechanisms are often inflexible and lack the expressiveness required for complex research workflows. OR-Agent adopts a more flexible and explicit approach, allowing the exploration-exploitation trade-off to be controlled at multiple levels of the research process. Specifically, this balance is governed through several complementary design dimensions:
\begin{itemize}
    \item Research tree shape control;
    \item Dynamic allocation of experimentation resources;
    \item Strategy of selecting parent heuristics from database.
\end{itemize}

Empirically, relatively simple optimization tasks can often be solved effectively by emphasizing evolutionary operations such as crossover and mutation, whereas more challenging problems require a more carefully calibrated balance between broad exploration of diverse research directions and intensive exploitation of promising candidates.

\section{Experiments}

We conduct experiments across 12 classical combinatorial optimization problems and simulation-based cooperative driving scenarios. All test scripts, hyperparameters, and full logs are available in the repository \cite{liu2026oragent} for reproducibility.

\subsection{Experimental Setup}

\paragraph{Benchmarks.}
The benchmark suite is adapted from ReEvo~\cite{ye2024reevo} and extended with additional problems, including a cooperative driving task implemented in SUMO. The suite includes 12 classical combinatorial optimization problems spanning routing (TSP, CVRP), packing (BPP, MKP), electronic design (DPP), orienteering (OP), and simulation-based multi-vehicle coordination.

\paragraph{Baselines.}
We compare OR-Agent against representative LLM-based algorithm discovery frameworks: \textit{FunSearch}~\cite{romera2024mathematical}, \textit{AEL}~\cite{liu2023algorithm}, \textit{EoH}~\cite{liu2024evolution}, and \textit{ReEvo}~\cite{ye2024reevo}. For the cooperative driving problem, we additionally compare against the SUMO~\cite{sumo2026} default driving model.

\paragraph{LLM Configuration.}
Two LLMs were evaluated: \textit{Qwen3}~\cite{yang2025qwen3} for the 12 classical OR problems, and \textit{DeepSeek V3.2}~\cite{liu2025deepseek} for the cooperative driving problem.

\paragraph{Performance Metric.}
To enable fair comparison across heterogeneous problems, we report a \emph{normalized score}. For problem $i$ and algorithm $j$, the normalized score is defined as:
$
\text{NormalizedScore}^i_j 
=
\frac{\left| \text{score}^i_j - \text{worst}^i \right|}
{\left| \text{best}^i - \text{worst}^i \right|}
$.

\paragraph{Computation Budget.}
Performance is evaluated under two complementary computational budgets: number of LLM calls, and number of function evaluations. Depending on the application scenario, either LLM inference cost or evaluation cost may become the primary bottleneck.

\subsection{Experiment Results}

\begin{figure}
    \centering
    \includegraphics[width=1.0\linewidth]{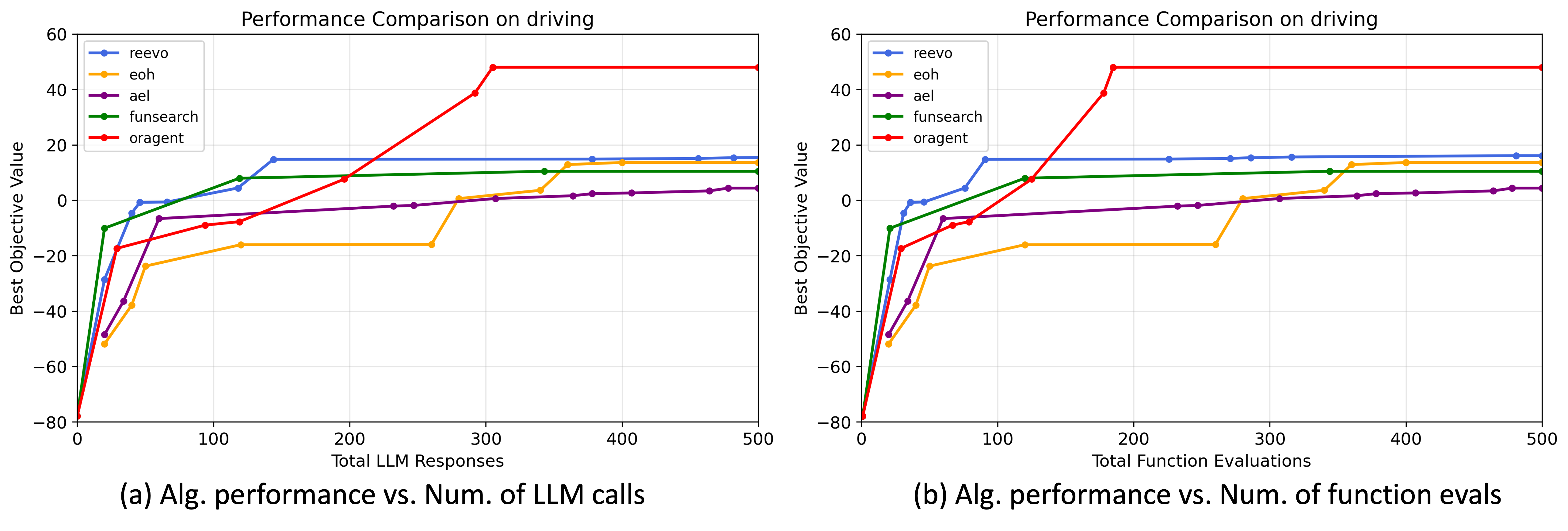}
    \caption{Performance comparisons on driving problem.}
    \label{fig:performance_comparision_driving}
\end{figure}

Table~\ref{tab:or_results} shows the normalized scores across 12 benchmark problems. OR-Agent achieves the highest average normalized score (0.924), substantially outperforming all baselines. Figure~\ref{fig:performance_comparision_driving} presents driving performance comparisons under equal computational budgets. Under constrained runtime, OR-Agent achieves a score of 48.00, significantly surpassing the highest baseline score (16.10), as shown by Table~\ref{tab:driving_detailed}. The gap becomes more pronounced as runtime increases. Notably, OR-Agent is the only algorithm that discovers a heuristic cooperative driving algorithm exceeding the SUMO default driving model after extended running, achieving an average score of 90.24 compared to 85.25.

\begin{table}
\centering
\caption{Normalized scores across 12 classical OR benchmark problems.}
\label{tab:or_results}
{\small
\setlength{\tabcolsep}{3pt}
\begin{tabular}{lccccc}
\toprule
Problem & FunSearch & AEL & EoH & ReEvo & OR-Agent \\
\midrule
TSP-Constructive & 0.000 & 0.878 & 0.788 & \textbf{1.000} & 0.959 \\
TSP-ACO          & 0.252 & 0.000 & 0.386 & 0.258 & \textbf{1.000} \\
TSP-POMO         & 0.000 & \textbf{0.975} & \textbf{1.000} & 0.771 & 0.884 \\
TSP-LEHD         & 0.648 & 0.000 & 0.175 & 0.259 & \textbf{1.000} \\
CVRP-ACO         & 0.000 & 0.403 & \textbf{1.000} & 0.471 & 0.680 \\
CVRP-POMO        & 0.000 & \textbf{1.000} & 0.418 & 0.681 & 0.986 \\
CVRP-LEHD        & 0.000 & 0.285 & 0.855 & 0.033 & \textbf{1.000} \\
BPP-Online       & 0.913 & \textbf{1.000} & 0.728 & 0.000 & 0.948 \\
BPP-Offline-ACO  & 0.290 & 0.270 & 0.000 & 0.290 & \textbf{1.000} \\
DPP-GA           & 0.294 & 0.000 & \textbf{1.000} & 0.190 & 0.787 \\
MKP-ACO          & 0.833 & 0.000 & 0.633 & 0.951 & \textbf{1.000} \\
OP-ACO           & 0.653 & 0.916 & 0.000 & \textbf{1.000} & 0.839 \\
\midrule
\textbf{Average} & 0.323 & 0.477 & 0.582 & 0.492 & \textbf{0.924} \\
\bottomrule
\end{tabular}
}
\end{table}

\begin{table*}
\centering
\small 
\caption{Detailed performance comparison on the cooperative driving problem. ``Low'' and ``High'' denote the low and high traffic demand cases, respectively.
Safety metrics include collisions and critical TTC events. 
Efficiency is measured by average speed (m/s), and smoothness by speed variance.}
\label{tab:driving_detailed}
\setlength{\tabcolsep}{3pt} 
\begin{tabular}{llcccccc}
\toprule
\multirow{2}{*}{Model} & \multirow{2}{*}{Case} 
& Collisions & Critital TTC & Avg Speed & Speed Var. & Case & Avg \\
\cmidrule(lr){3-8}
 &  &  &  & (m/s) &  & Score & Score \\
\midrule
\multirow{2}{*}{SUMO Default} 
& Low & 0 & 10 & 11.63 & 6.48 & 95.18 & \multirow{2}{*}{85.25} \\
& High & 0 & 60 & 10.36 & 11.00 & 75.32 &  \\
\midrule
\multirow{2}{*}{OR-Agent} 
& Low & 0 & 12 & 11.94 & 5.31 & 96.08 & \multirow{2}{*}{\textbf{90.24}} \\
& High & 0 & 46 & 11.12 & 7.74 & 84.40 &  \\
\bottomrule
\end{tabular}
\end{table*}

\subsection{Ablations}

We conduct ablation studies along two dimensions: thinking depth and memory compression. All ablations are performed under fixed computational budgets to ensure fair comparison.

\begin{figure}
    \centering
    \includegraphics[width=0.80\linewidth]{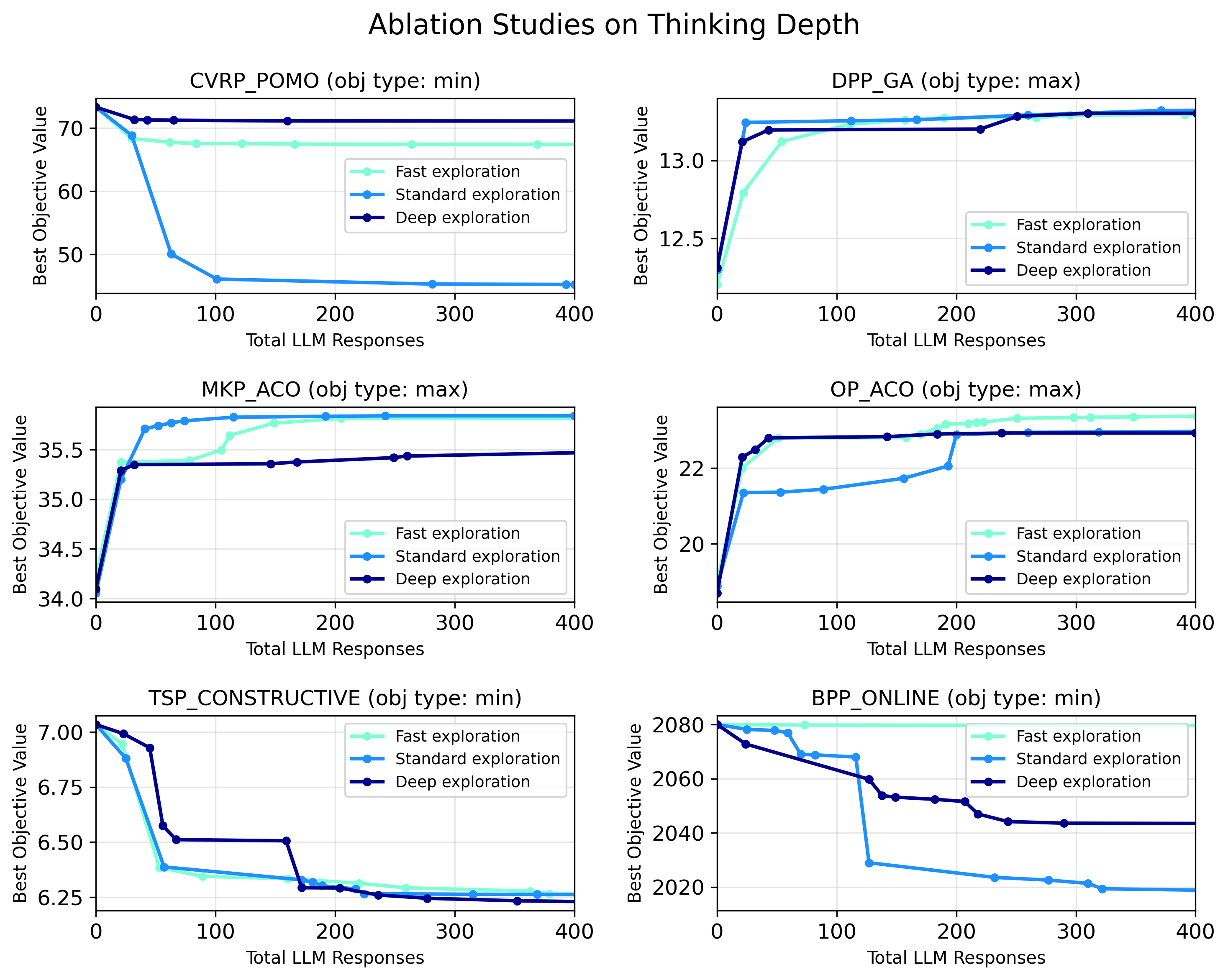}
    \caption{Ablation on thinking depth. Balanced tree depth yields superior performance compared to both excessively deep and overly shallow configurations.}
    \label{fig:ablation_thinking_depth}
\end{figure}

\paragraph{Ablation on Thinking Depth.}
We investigate the classical exploration–exploitation trade-off by varying the depth and per-node experimentation budget of the research tree. Under a fixed total computational budget, deeper investigation of specific regions necessarily reduces overall coverage of the search space. We compare three configurations: ``deep exploration'' mode with increased maximum tree depth and per-node experiment limit; ``fast exploration'' mode with reduced depth and limited refinement iterations; and ``standard'' mode with intermediate parameters (default setting). As shown in Figure~\ref{fig:ablation_thinking_depth}, both ``deep exploration'' and ``fast exploration'' underperform the standard configuration. Excessively deep search limits diversity and increases the risk of over-investing in suboptimal regions, while overly shallow search sacrifices refinement quality.

\begin{figure}
    \centering
    \includegraphics[width=0.80\linewidth]{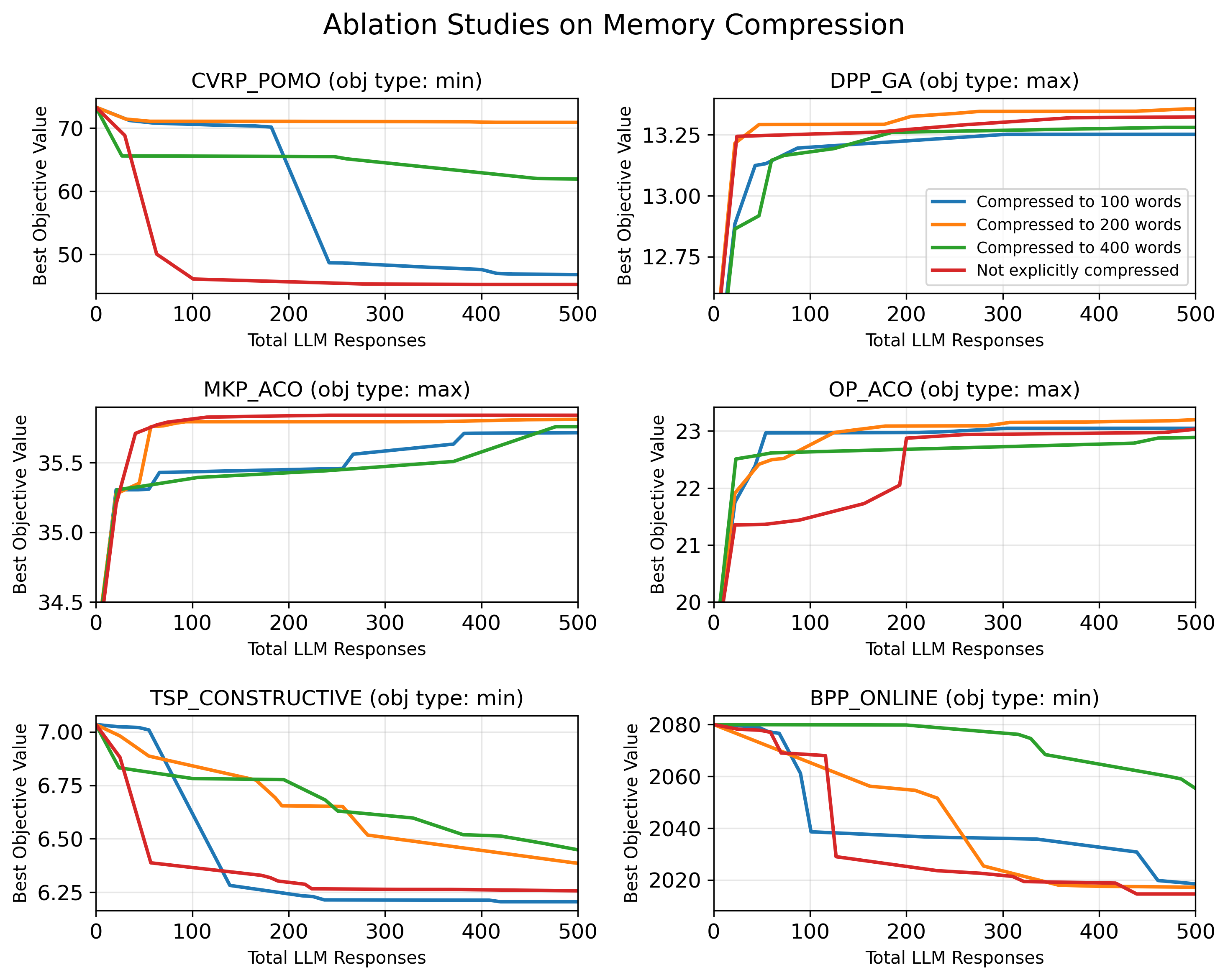}
    \caption{Ablation on long-term reflection compression. No compression achieves the strongest overall performance, while moderate compression performs worst.}
    \label{fig:ablation_memory_compression}
\end{figure}

\paragraph{Ablation on Memory Compression.}
Long-term reflection serves as accumulated verbal momentum. Compressing reflection memory functions analogously to gradient decay in optimization, potentially attenuating accumulated directional signals. We evaluate four compression levels: 100 words, 200 words, 400 words, and no explicit compression (approximately 800 words under the tested models) (Figure \ref{fig:ablation_memory_compression}). Across six representative problems, no explicit compression achieves the best performance in four cases and yields the highest overall average score. Interestingly, aggressive compression to 100 words performs second-best, while moderate compression (400 words) performs worst. This pattern suggests that partial compression may remove useful high-level signals without sufficiently improving signal-to-noise ratio. The exact mechanisms underlying this phenomenon warrant further investigation.

\section{Related Work}

AI techniques for operations research have evolved from reinforcement learning and neural combinatorial optimization approaches \cite{zhang1995reinforcement} to LLM-based methods that generate solutions, improve quality through iterative prompting~\cite{yang2023large}. Hyper-heuristic methods provide a unifying perspective by searching over heuristic spaces \cite{romera2024mathematical,liu2023algorithm,liu2024evolution,ye2024reevo}. However, most existing systems remain limited to simple evaluation settings and lack mechanisms for deep, environment-driven exploration, which OR-Agent addresses by explicitly targeting complex environments requiring iterative experimentation.

Building on traditional genetic programming \cite{koza1994genetic}, recent work has leveraged LLMs as flexible evolutionary operators \cite{lehman2023evolution}. FunSearch \cite{romera2024mathematical,liu2023algorithm} demonstrated LLM-guided mutation for scientific discoveries. Subsequent systems including EoH \cite{liu2024evolution}, ReEvo \cite{ye2024reevo}, and AlphaEvolve \cite{novikov2025alphaevolve} further explored reflective mechanisms and code-level evolution. Despite progress, these methods often rely on shallow fitness feedback without systematic refinement. OR-Agent extends this work by embedding evolutionary operators within a tree-search workflow that supports deep iterative refinement, explicit environment exploration, and reflection mechanisms that reinterpret optimization concepts at the level of natural language and code.

\section{Conclusion}

We presented OR-Agent, a multi-agent framework for automated algorithm discovery that unifies evolutionary search, tree-structured research workflows, hierarchical reflection mechanisms, and a structured heuristics database. Rather than relying solely on mutation and crossover, OR-Agent models heuristic design as a structured process of branching hypothesis exploration, systematic refinement, iterative experimentation, and knowledge accumulation. Its hierarchical reflection system serves as a semantic analogue of optimization dynamics, guiding improvement across research rounds. Extensive experiments on twelve classical combinatorial optimization problems and one complex cooperative driving environment show that OR-Agent achieves the strongest overall performance among state-of-the-art LLM-based automatic heuristic search methods. For future research, extending beyond purely textual feedback to incorporate multi-modal signals would enhance diagnostic and reasoning capabilities. Equipping the Idea Agent with web search and literature review capabilities could further enhance its ability to generate innovative ideas.

\section{Acknowledgment}
This research was funded by the National Key Research and Development Program of China (grant number 2024YFB4303100, 2024YFB4303104-02), and Shanghai Baiyulan Talent Project Pujiang Program (grant number 24PJD115).



%
%
%
%
\bibliographystyle{splncs04}
\bibliography{references}

\end{document}